# Follow-Up Questions Improve Documents Generated by Large Language Models

Bernadette J Tix[1]

## Abstract

This study investigates the impact of Large Language Models (LLMs) generating follow-up questions in response to user requests for short (1-page) text documents. Users interacted with a novel web-based AI system designed to ask follow-up questions. Users requested documents they would like the AI to produce. The AI then generated follow-up questions to clarify the user's needs or offer additional insights before generating the requested documents. After answering the questions, users were shown a document generated using both the initial request and the questions and answers, and a document generated using only the initial request. Users indicated which document they preferred and gave feedback about their experience with the question-answering process. The findings of this study show clear benefits to question-asking both in document preference and in the qualitative user experience. This study further shows that users found more value in questions which were thought-provoking, open-ended, or offered unique insights into the user's request as opposed to simple information-gathering questions.

## Introduction

Advances in generative AI have made it possible for a software program to produce a broad range of useful output from natural language prompts, including visual artwork [1], music [2], working software code [3, 4], and text [5].

It has long been known that ambiguity is pervasive within natural language. This ambiguity has historically caused difficulties in parsing any natural language request into actionable software output [6–9]. This pervasive ambiguity can cause confusion in human communication as well. When developing new software, for example, an active and involved process of requirements gathering is typically necessary before the development team can begin their work [10]. One of the most straightforward ways of overcoming miscommunications is simply to ask questions [11, 12]. However, the most widely available Larg Language Models (LLMs) do not, by default, ask follow-up questions in response to confusing or ambiguous prompts. Instead, publicly available models including ChatGPT [13], Gemini [14], and Bing [15] will attempt to fulfil requests from the user with whatever information they have been given.

This is not to say that LLMs are incapable of generating useful questions. When specifically prompted to do so, LLMs can generate relevant questions and produce improved output in response to those questions [16–18]. Previous work in this area has focused on disambiguation for short questions [16] and simple task requests [18], with the primary goal of achieving complete disambiguation of the user request [17]. Prior work has also shown that LLMs can produce more useful output when given access to the full conversation that led up to the user query [19].

---

[1] University of Hawai'i at Mānoa (*https://orcid.org/0009-0005-3460-7657*) *bjavery@hawaii.edu*

Questions and dialog in response to a user prompt are capable of much more than disambiguation. LLMs can be trained to ask questions in educational settings to act as a tutor [20, 21] and to generate assessment questions for teachers [22]. LLMs are also capable of assisting with organizing, outlining, and other tasks related to writing [23] and can even improve critical thinking skills when employed properly [24].

This paper investigates whether there is value in an LLM asking follow-up questions when prompted to produce a short document such as a letter, memo, email, or a short report. This task is inherently more ambiguous than answering a short question or performing a simple task, as there is no single "correct" document. Participants in this study prompted a generative AI to produce a document they desired, answered AI-generated follow-up questions about their needs, and then compared and rated a pair of documents. One document was generated taking the user's questions and answers into consideration, and the other document was generated based on only the user's original request. Participants also gave qualitative feedback about the experience and answered an exit survey targeted at determining whether there was value in the question-answering process itself, apart from final document preference.

## Materials and Methods

This study was carried out using a web-based application called the Clarifying Questions Document Generator (CQDG) created specifically for this research. The key components of CQDG are:

- A user-facing front-end.
- A back-end powered by three different LLMs, including:
    - GPT-3.5 Turbo by OpenAI [25]
    - GPT-4.0 Turbo by OpenAI [26]
    - Gemini Pro by Google [14]
- A database for logging results from the use of the system, implemented using Microsoft Azure Data Services [27]

### User Experience

CQDG begins by presenting users with an informed consent. Once the user has read and agreed to the consent, they are asked to report their age, gender, level of prior experience with AI, and whether they are fluent in English. Users who are under 18 or are not fluent in English are informed they are not eligible to participate. Users who are over 18 and fluent in English are presented with the following prompt:

*"On this page, you will be communicating with an AI that is capable of writing short documents such as letters, memos, emails, and short reports. Please think of a document you would like the AI to create for you. This could be a document you actually need, or one that you have just made up for*

*the experiment. Either way, please think in detail about what you would need this document to include."*

A text-input box is provided for users to enter a prompt describing the document they would like to create. CQDG then generates three follow-up questions intended to gather additional context or get the user to think about their request in ways they may not have previously considered. These questions are presented to the user, and they are prompted to answer each question before continuing. CQDG then generates two documents. One document uses both the user's original prompt and the subsequent questions and answers as context for the document generation (*QA document*). The other document uses only the user's original prompt and disregards the questions and answers (*baseline document*). The two documents are presented to the user side-by-side. The order of the documents is randomized, so half of the users see the baseline document on the left and the QA document on the right, while the other half see the documents in the opposite order. Beneath the documents, two questions are presented with sliders allowing the user to select which document they prefer. The two questions are:

- *Which document do you prefer overall?*
- *Which document would be more useful to you in its current state?*

After rating the documents, users are given the opportunity to continue refining one or both documents with additional instructions, up to three times per document. Refining is optional, but if the user chooses to refine at least one document, they are prompted to give new ratings to the same questions after refining is complete.

After giving their final ratings, users are given the option to create a new document or proceed to the exit survey. Those that choose to create a new document are taken back to the screen where CQDG asked for an initial document prompt and the study continues as before. Users do not have to redo the consent, screener, or demographic questions when making a new document. Users can continue creating as many documents as they would like before proceeding to the exit survey.

The exit survey includes five statements, with sliders allowing users to indicate their agreement or disagreement with each statement on a scale of 0-10:

- *It was annoying to have to answer questions even though I had already explained what I wanted the AI to do.*
- *I felt like the AI was more engaged with my problem because it asked follow-up questions.*
- *I would be willing to answer follow-up questions from an AI if answering questions led to better results.*
- *I liked that the AI showed me two options to pick between, instead of only picking the option it thought was best.*
- *Answering the questions asked by the AI made me think about my request in ways I hadn't previously considered.*

Users are also given the opportunity to provide free-text feedback at several points in the program, including when answering CQDG-generated questions, when reviewing the initial or refined documents, and at the exit survey.

# Technical Design

The frontend of CQDG was kept intentionally simple, run from a single HTML page, with all functionality contained within embedded JavaScript. The webpage is able to interact with various LLM APIs as well as with an Azure Database by calling on Serverless Azure Functions [28].

When a user navigates to the CQDG webpage, hosted as a GitHub site [29], CQDG establishes a connection to the database and randomly selects one of the three available LLMs for this study. The selection of LLM between GPT 3.5, GPT 4, and Gemini is invisible to the user. Whichever LLM is selected will be used throughout document creation, question generation, and document refining. If the user opts to create a new document, the LLM random selection process will be re-run when the user enters their new prompt. The software design is shown in Figure 1.

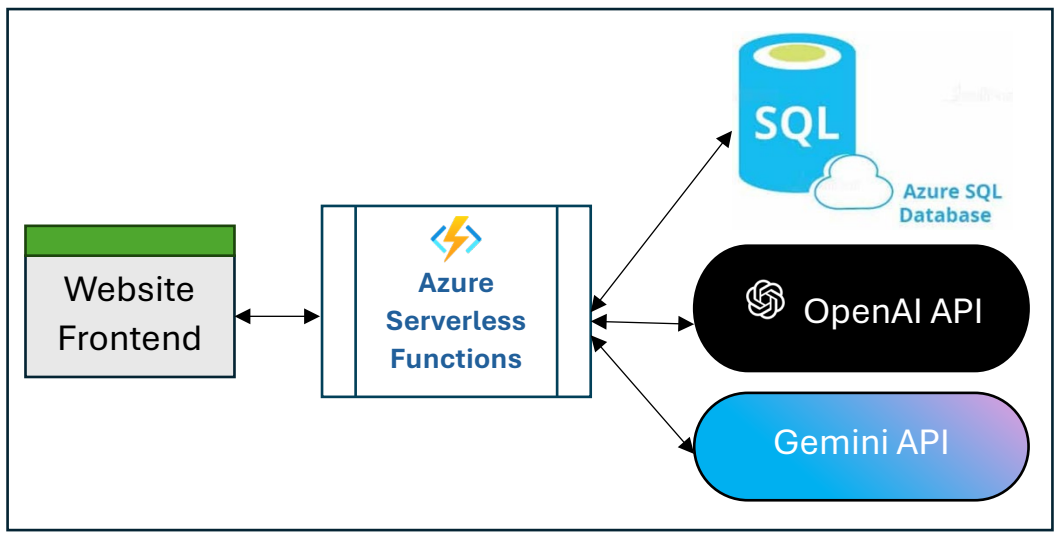

Figure 1: CQDG Software Modules Diagram

When the user enters their initial document request prompt, a two-step process is used to generate questions. First, CQDG prompts the LLM to identify areas that are potentially promising to ask questions about, using the following prompt:

---

*A user is requesting the creation of a new document. This is their request:*
*user: "<prompt>"*
*Identify any areas of significant ambiguity in the prompt, areas that could benefit from*
*more thought or attention from the user, or helpful tips the user may not have considered.*
*Write these out in a short list.*

---

Where *<Prompt>* is the prompt entered by the user. After the LLM has given a response to this, the questions are generated with a second call to the LLM:

> *Pick the three most important items from the list you just generated, and write a list of three insightful questions that will improve the requested document. Phrase the questions as direct questions to the user. Format your response as a numbered list of exactly 3 questions.*

This reliably produces a list of three numbered questions, which are then parsed and presented to the user separately, each with their own text area input for the user to provide their answers. Once the answers are submitted, the questions and answers are re-arranged to produce an artificial conversation history, making it appear as though the conversation progressed naturally in the following order: *initial user prompt, first question, first answer, second question, second answer, third question, third answer.* This creates a natural-appearing conversation for the LLM to produce a continuation from but does not provide clear instruction to the LLM on what to produce next. Thus, an additional assistant message and user message are appended to the end of the conversation:

> *Assistant: Thank you for your answers. I will now create a document based on the questions and answers you have provided. Do you have any further instructions?*
>
> *User: Generate a high-quality document that meets the user's needs, considering both their initial prompt and the answers they gave when asked for details. Include creative original insights that will improve the quality of the document but do not deviate too far from the user's original intent.*

This full conversation is then sent to the LLM, and is used to produce the QA document. The baseline document is produced by sending only the original user prompt but none of the questions or answers. All conversations are prepended with a system-level instruction[2] to guide the tone of the output. For all conversations sent in the QA process, this message is:

> *You are a helpful assistant designed to help users create short, high-quality documents by asking insightful questions to clarify the user's needs and make them think about things they have not considered, and then create high-quality professional documents after discussing the details with the user.*

For the Baseline document, the system message is simpler, and excludes the statement about asking questions:

---

[2] The OpenAI API allows the use of system-level messages, but Gemini does not. When sending conversations to Gemini, system messages are sent as user messages.

> *You are a helpful assistant designed to help users create short, high-quality professional documents.*

When refining the documents, the LLM is sent the entire conversation that led to the creation of the document being refined, including the document itself, followed by an additional prompt:

> *The user has provided some additional feedback. Please re-write the entire document, modifying the original based on this new feedback: "<feedback>"*

Where *<feedback>* is the refining prompt entered by the user. The same message format is used when refining either the QA or Baseline document. A flowchart showing each step of this process is provided in Figure 2.

## Design Improvements from Pilot Study

This study was informed by a pilot study conducted in January 2024 [30]. Several key improvements were made to CQDG based on the insights from the pilot study. These improvements include:

- Question generation prompts were refined to encourage the generation of insightful or thought-provoking questions. During the pilot study, some of the questions generated by CQDG were simple fill-in-the-blank questions such as asking for the user's name or the name of their organization, which users did not find valuable as this is information that could easily have been added in editing rather than through a question-and-answer dialog.
- Document generation for the QA document was refined to emphasize retaining a high degree of originality and creativity. A common complaint of users in the pilot study was that the QA documents stuck too closely to the information they had been given, whereas the baseline documents often included original insights.

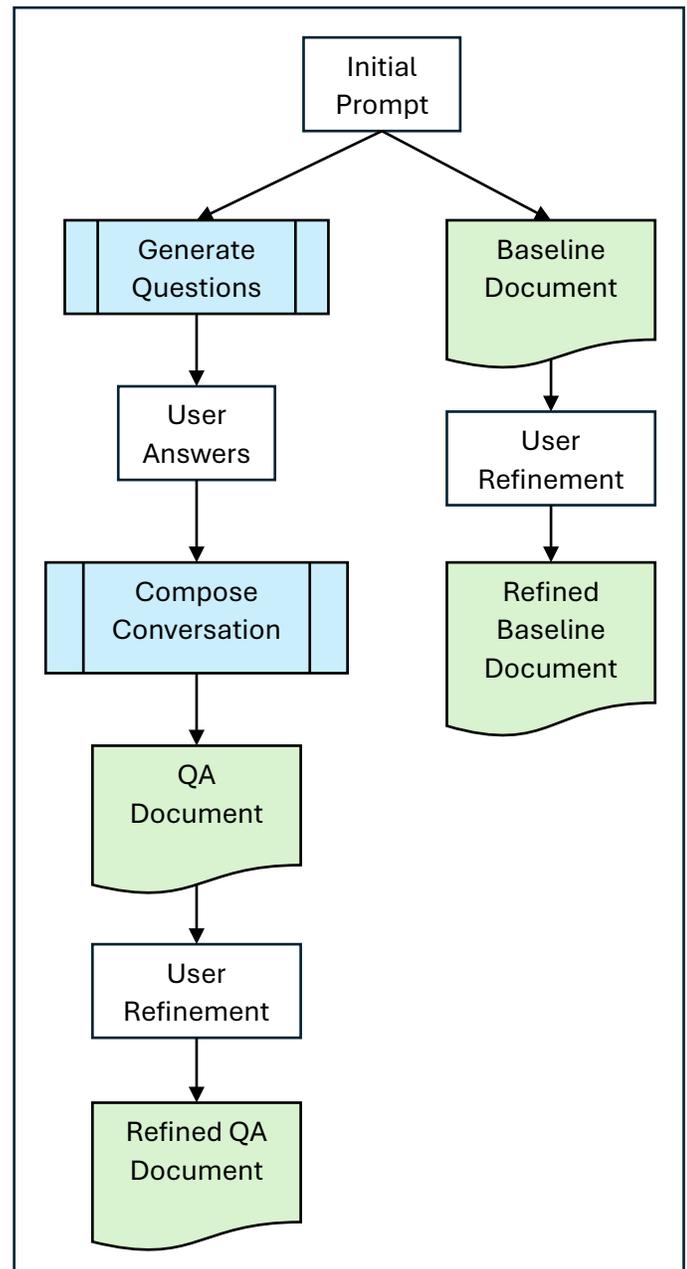

*Figure 2: Document Generation Flowchart*

- Documents were displayed side by side and users were asked for their preference between the two documents. In the pilot study, documents were presented one after the other on separate screens and users were asked to rate each document independently.
- Users were given the option to refine each document after its initial creation. This was the most-requested feature from the pilot study.
- CQDG was adapted to include GPT-3.5 Turbo, GPT-4 Turbo and Gemini Pro LLMs. The pilot study only included GPT-3.5 Turbo.

# Results

Study participants were invited to take the study via a social media ad campaign that directed users to the CQDG website. Users entered a total of 89 prompts into CQDG, but 14 exited before completing the study, leaving 75 who completed. Incomplete studies were discarded. Of the completed responses, four were excluded from the study. The excluded responses include a request for an obscene document which Gemini refused to produce, a user who stated that they did not want CQDG to produce anything and reiterated that they did not want any output at the question-answering phase, and two technical glitches in which one of the documents failed to generate.

The 71 remaining responses were submitted by 65 unique respondents. 25 respondents reported being male, 33 female, and 7 selected "other / nonbinary" for their gender. The youngest respondent was 18[3] and the oldest was 64, with an average age of 31. Six users chose to stay in CQDG and create an additional document after completing their first document. 5 users reported having never used generative AI before, 35 said they had used generative AI before, but not often, and 25 reported being regular users of generative AI. In 36 of the responses, users chose to refine one or both documents with additional instructions after giving their initial ratings.

The tone of requested documents varied widely. Some were light and silly, such as *"Send my cat a divorce letter."* Others were serious, such as *"Write an email to a friend I haven't spoken to in 20 years because of a fight saying I want to repair the relationship."* Others were detailed and professional, such as *"Please write a letter to parents of high school students who are members of a student robotics club. The letter should explain to parents that the club will be forming a board, and the club is looking for parent volunteers to serve on that board. The letter should be brief while maintaining a tone that is semi-formal and upbeat / enthusiastic. The letter should set a specific date for the initial board meeting and encourage parents to attend this meeting if they are interested in serving or would like to know more about club management and future plans."*

Users were asked two questions about the produced documents: *"Which document do you prefer overall?"* and *"Which document would be more useful to you in its current state?"* These questions were asked both before and after users were given the opportunity to refine both documents with additional instructions. Table 1 shows the responses to these questions.

---

[3] Respondents who reported being younger than 18 were screened out and not allowed to continue with the study.

*Table 1: Preference Results*

|  | Before Refining | | | After Refining | | |
|---|---|---|---|---|---|---|
|  | **Prefer QA Document** | **Prefer Baseline** | **No Preference** | **Prefer QA Document** | **Prefer Baseline** | **No Preference** |
| **Overall Preference** | 41 | 25 | 5 | 18 | 13 | 5 |
| **More Useful** | 40 | 22 | 9 | 13 | 12 | 11 |

Figure 3 shows the distribution of answers for each of the 3 LLMs used. GPT-4 consistently resulted in a greater user preference for the QA document over the baseline compared to GPT-3.5 or Gemini, which showed comparatively more equal preference for the baseline & QA documents. This is particularly true after refining the documents. After refining, user preference for the baseline or the QA document is equal or nearly equal for documents created using GPT-3.5 or Gemini, but a strong preference for the QA document is retained after refining for documents created with GPT-4.

The preference data was tested for significance using a binomial distribution test. Since the binomial distribution test requires that there be only two possible categories, so "No Preference" responses were distributed equally between "QA" and "Baseline" responses. This can reduce the sensitivity of the tests, but is less likely to skew the results than either proportional distribution or excluding the "no preference" responses [31]. Furthermore, in this study a response of "no preference" represents equal preference for either option, thus distributing the "no preference" responses equally between the available options is more appropriate than discarding these responses. Since the binomial distribution test requires a whole number of responses, in cases where there were an odd number of "no preference" responses, one "no preference" response was discarded to avoid fractional values.

*Table 2: Test of Significance*

| *p values under Binomial Distribution Test* | **Before Refining** | **After Refining** |
|---|---|---|
| **Overall Preference** | *0.028* | *0.155* |
| **More Useful** | *0.016* | *0.368* |

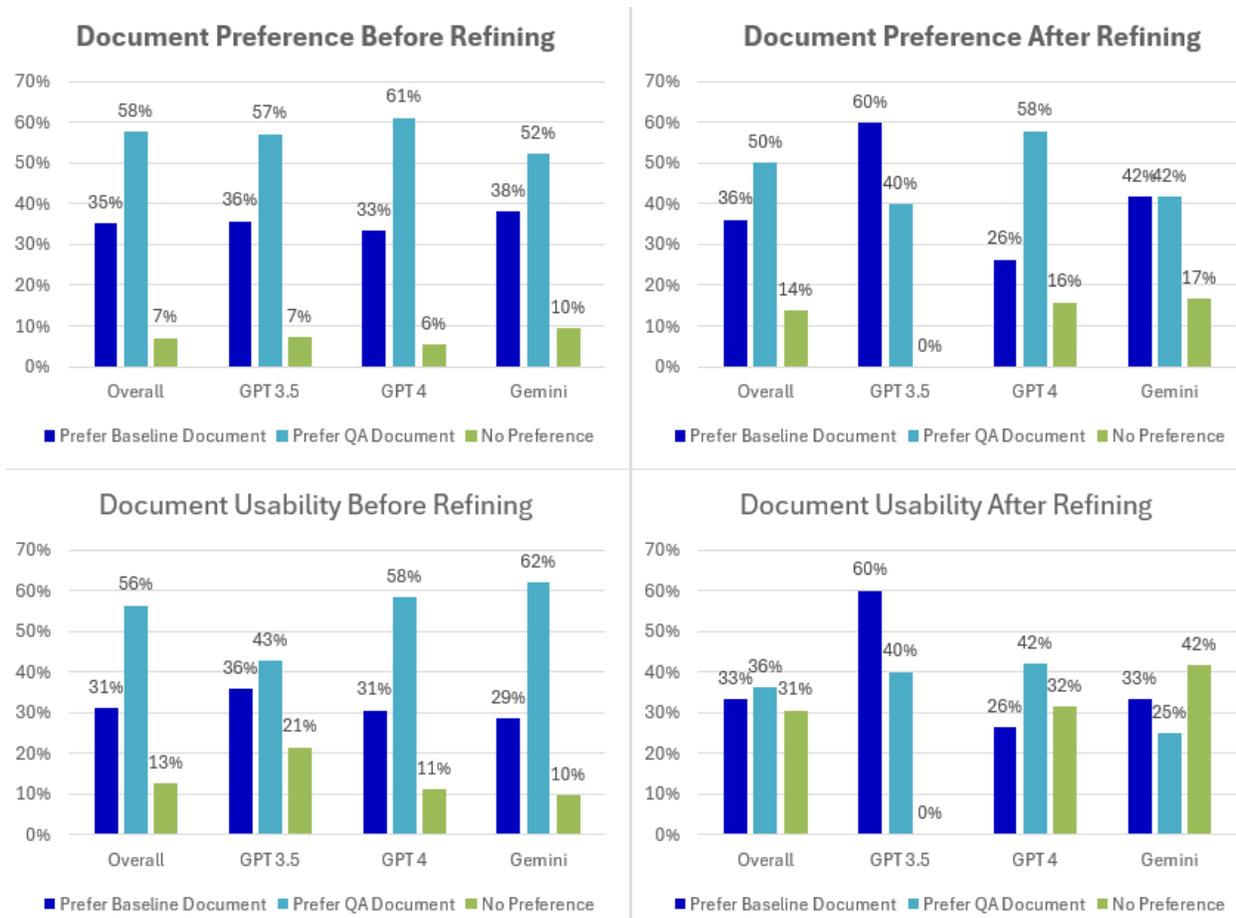

*Figure 3: Impact of LLM on User Preferences*

Women showed an overall greater preference for the QA document than men, as shown in Figure 4. However, this difference was not found to be significant under a 2-sample T-test comparing men's and women's responses. Users who selected "other/nonbinary" were excluded from the gender comparison due to the low number of users who selected this option.

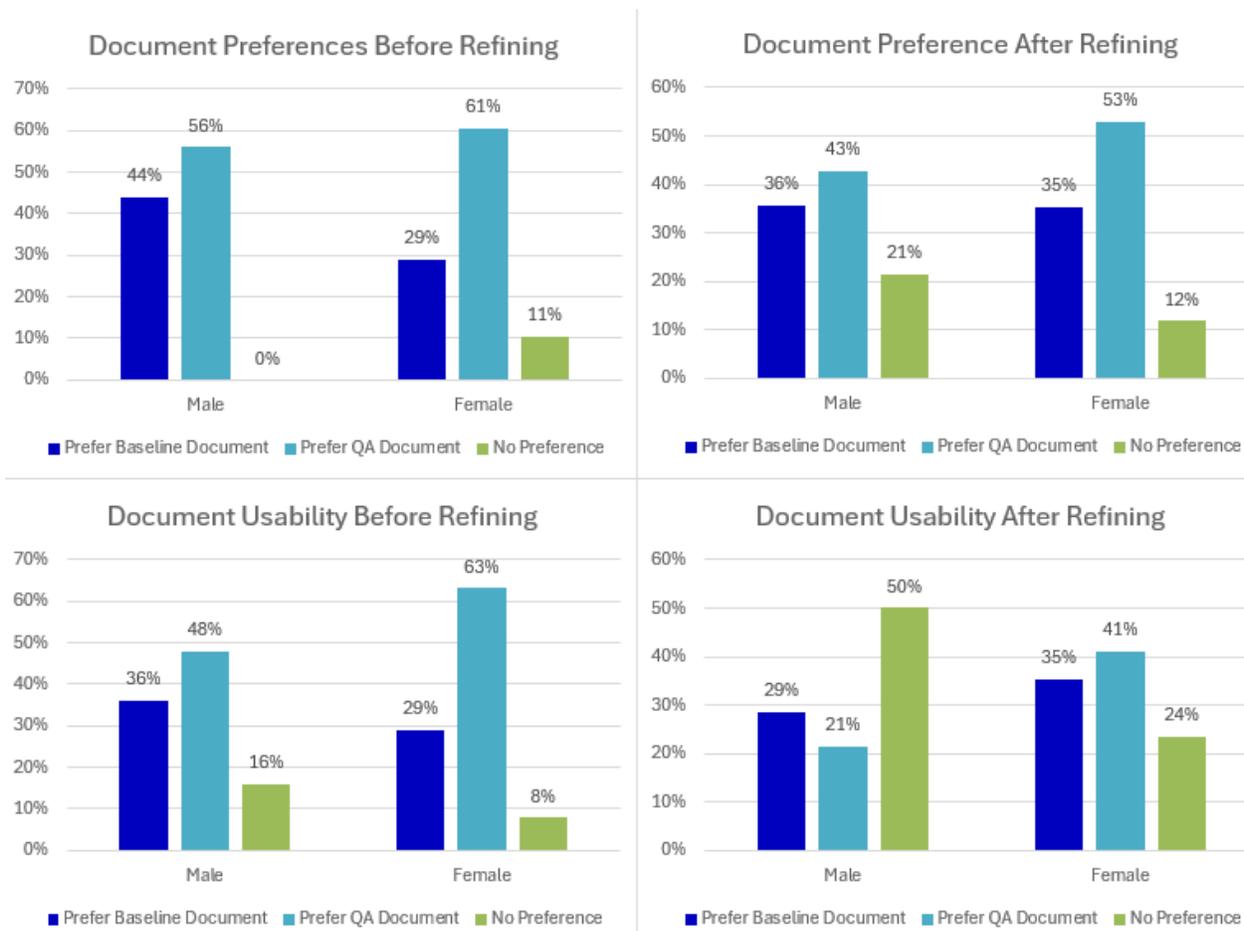

*Figure 4: Impact of Gender on User Preferences*

After users were finished creating and refining as many documents as they desired, they were given an exit survey with five questions. 65 exit surveys were collected, one per unique user. Exit survey results are shown in Figure 5. Most users disagreed with the statement "*It was annoying to have to answer questions even though I had already explained what I wanted the AI to do,*" (59% disagree). Most agreed with the statements *"I felt like the AI was more engaged with my problem because it asked follow-up questions,"* (64% agree), *"I would be willing to answer follow-up questions from an AI if answering questions led to better results,"* (84% agree), *"I liked that the AI showed me two options to pick between, instead of only picking the option it thought was best,"* (78% agree) and *"Answering the questions asked by the AI made me think about my request in ways I hadn't previously considered"* (74% agree).

Additionally, users were able to enter free-text feedback at six points in the study: after answering questions but before seeing the initial documents, once for each of the initial two documents, once for each of the two refined documents, and once after filling out the exit survey. The top 5 repeating themes in these free-text responses have been identified and listed in Table 3.

Overall, users were impressed with the quality of the questions being asked and found the process of considering and answering the questions to be thought-provoking and a valuable step in document creation. One user stated, *"I would find answering these question prompts valuable, even if I were still writing the letter myself."* Another noted that the questions themselves might be relevant to include in the document they were trying to create. A minority of users expressed frustration at the question-answering process. Many users expressed finding the experience of creating documents with CQDG to be engaging and noted that the experience increased their level of interest in and understanding of AI. Several users stated that after the refining process, the two documents improved in quality and were more similar in quality than they had been before refining. This feedback agrees with the preference analysis described above, in which the preference for the QA document is significant before the refining process, but the difference becomes insignificant after refining. However, several users noted frustration with the refining process and an inability to get the document to come out the way they envisioned it, while noting that the initial QA document made a better starting point for manual revisions when compared to the initial baseline document.

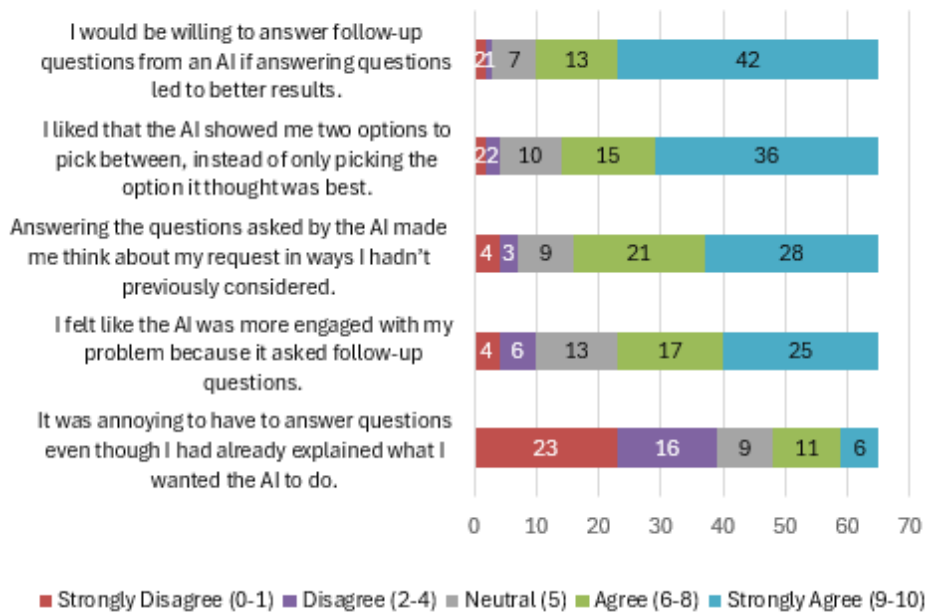

*Figure 5: Exit Survey Responses*

*Table 3: Top 5 Themes for each Category of Free-Entry Feedback*

| Category | Themes |
|---|---|
| After Answering Questions | <ul><li>The AI came up with good questions.</li><li>Reading and answering the questions was valuable in itself.</li><li>Answering the questions was thought-provoking.</li><li>Users offering suggestions for other questions the AI could have asked.</li><li>Users attempting to provide additional instructions to the AI using the feedback box.</li></ul> |
| Initial Baseline Document | <ul><li>The document was of a high quality.</li><li>The content was vague or generic.</li><li>The language used was too polite or too flowery.</li><li>Users noted hallucinations in the output.</li><li>The document's tone was awkward or sounded artificial.</li></ul> |
| Initial QA Document | <ul><li>This document is better than the other (baseline) document.</li><li>The document was of a high quality.</li><li>The document is too long or too wordy.</li><li>The AI failed to follow specific instructions given by the user such as length limits.</li><li>The document is a good starting point for the user to edit into a final document.</li></ul> |
| Refined Baseline Document | <ul><li>Revision improved the document.</li><li>The document was of a high quality.</li><li>Refining failed to fix the problems with the document / AI failed to respond to revision prompts the way I would hope.</li><li>Users noted hallucinations in the output.</li><li>This document could be used as-is.</li></ul> |
| Refined QA Document | <ul><li>Revision improved the document.</li><li>The document was of a high quality.</li><li>Refining failed to fix the problems with the document / AI failed to respond to revision prompts the way I would hope.</li><li>The AI failed to follow specific instructions given by the user such as length limits.</li><li>This document could be used as-is.</li></ul> |
| Exit Survey | <ul><li>The study was enjoyable.</li><li>Answering questions was valuable to the user and an improvement over the usual process of generating documents with AI.</li><li>Users expressing increased interest in AI after participating in the study.</li><li>Users had a positive experience interacting with CQDG.</li><li>*(Only 4 themes identified due to lower response rate for optional exit survey feedback).*</li></ul> |

## Discussion

The key finding of this study is that users found value in generative AI that asks open-ended, thought-provoking questions before producing the requested output. Establishing a cooperative dialog between users and generative AI, rather than a simple request fulfilment model, has the potential to enhance the user experience, engage users more deeply with the problems they are trying to solve, and produce higher-quality generated documents. Of the three LLMs tested, this effect was strongest with GPT-4 and weakest with GPT-3.5, suggesting that asking and responding to follow-up questions is more effective with higher-quality LLMs.

Users also responded positively to working with the system, with many users stating that they found answering questions to be valuable, engaging, and even enjoyable. Although similar levels of document quality can be achieved through successive refining prompts, follow-up questions are a promising alternative that may offer a better user experience as well as prompting users to engage thoughtfully with their problems in new ways.

LLMs have given modern software a capability that has long been out of reach, the ability to translate a natural language request into useful machine output. However, this brings with it all the problems and issues that natural communication between humans has always had, such as ambiguity, lack of context, and misunderstanding. Asking follow-up questions is a key tool in human communication, and one which future designs of AI-human interactive systems should continue to develop and utilize to the fullest.

## Declaration of Interest

There are no competing interests to declare. This work was self-funded by the author.

# Author Biographical Note

Bernadette Tix is a current PhD student at the University of Hawai'i at Mānoa, with 14 years of professional software development experience. Bernadette has a B.S. in Computer Science and a M.S. in Mechanical Engineering, with a background in robotics, automation, industrial control software, and enterprise IT.

# Data Availability Statement

All data used in this study was gathered independently. User responses were gathered under a confidentiality agreement and cannot be distributed in full. Aggregate data can be provided upon request.

# Human Research Ethics Declaration

This research was approved by the Institutional Review Board for the University of Hawai'i at Mānoa, Protocol ID: 2024-00193. All participants provided informed consent prior to participation.